# Comparative study of Financial Time Series Prediction By Artificial Neural Network with Gradient Descent Learning

Arka Ghosh

**Abstract**— Financial forecasting is an example of a signal processing problem which is challenging due to Small sizes, high noise, non-stationarity, and non-linearity,but fast forecasting of stock market price is very important for strategic business planning.Present study is aimed to develop a comparative predictive model with Feedforward Multilayer Artificial Neural Network & Recurrent Time Delay Neural Network for the Financial Timeseries Prediction.This study is developed with the help of historical stockprice dataset made available by GoogleFinance.To develop this prediction model Backpropagation method with Gradient Descent learning has been implemented.Finally the Neural Net ,learned with said algorithm is found to be skillful predictor for non-stationary noisy Financial Timeseries.

—————— ◆ ——————

## I. INTRODUCTION

Over past fifteen years, a view has emerged that computing based on models inspired by our understanding of the structure and function of the biological neural networks may hold the key to the success of solving intelligent tasks by machines like noisy time series prediction and more[1]. A neural network is a massively parallel distributed processor that has a natural propensity for storing experiential knowledge and making it available for use. It resembles the brain in two respects: Knowledge is acquired by the network through a learning process and interneuron connection strengths known as synaptic weights are used to store the knowledge[2]. Moreover, recently the Markets have become a more accessible investment tool, not only for strategic investors but for common people as well. Consequently they are not only related to macroeconomic parameters, but they influence everyday life in a more direct way. Therefore they constitute a mechanism which has important and direct social impacts. The characteristic that all Stock Markets have in common is the uncertainty, which is related with their short and long-term future state. This feature is undesirable for the investor but it is also unavoidable whenever the Stock Market is selected as the investment tool. The best that one can do is to try to reduce this uncertainty. Stock Market Prediction (or Forecasting) is one of the instruments in this process. We cannot exactly predict what will happen tomorrow, but from previous experiences we can roughly predict tomorrow. In this paper this knowledge based approach is taken.

The accuracy of the predictive system which is made by ANN can be tuned with help of different network architectures. Network is consists of input layer ,hidden layer & output layer of neuron, no of neurons per layer can be configured according to the needed result accuracy & throughput,there is no cut & bound rule for that.the network can be trained by using sample training data set,this neural network model is very much useful for mapping unknown functional dependencies between different input & output tuples.In this paper two types of neural network architecture,feed forward multilayer network & timedelay recurrent network is used for the prediction of the NASDAQ stock price.A comparative error study for both network architecture is introduced in this paper.

In this paper gradient descent backpropagation learning algorithm is used for supervised training of both network architectures. The back propagation algorithm was developed by Paul Werbos in 1974 and it is rediscovered independently by Rumelhart and Parker. In backpropagation learning atfirst the network weight is selected as random small value then the network output is calculated & it is compared with the desired output,difference between them is defined by error .The goal of efficient network training is to minimize this error by monotonically tuning the network weights by using gradient descent method.To compute the gradient of error surface it takes mathematical tools & it is a iterative process.

ANN is a powerful tool widely used in soft-computing techniques for forecasting stock price.The first stock forecasting approach was taken by White,1988 ,he used IBM daily stock price to predict the future stock value[3].When developing predictive model for forecasting Tokyo stock market , Kimoto, Asakawa, Yoda, and Takeoka 1990 have reported onthe effectiveness of alternative learning algorithms and prediction methods using ANN[4]. Chiang, Urban, and Baldridge 1996 have used ANN to forecast the end-of-year net asset value of mutual funds[5]. Trafalis (1999) used feed-forward ANN to forecast the change in the S&P(500) index. In that model, the input values were the univariate data consisting of weekly changes in 14 indicators[6].Forecasting of daily direction of change in the S&P(500) index is made by Choi, Lee, and Rhee 1995[7]. Despite the wide spread use of ANN in this domain, there are significant problems to be addressed. ANNs are data-driven model (White, 1989[8]; Ripley, 1993[9]; Cheng & Titterington, 1994[10]), and consequently, the underlying rules in the data are not always apparent (Zhang, Patuwo, & Hu, 1998[11]). Also, the buried noise and complex dimensionality of the stock market data makes it difficult to learn or re-estimate the ANN parameters (Kim & Han, 2000[12]). It is also difficult to come with an ANN architecture that can be used for all domains. In addition, ANN occasionally suffers from the overfitting problem (Romahi & Shen, 2000[13])[14].

## II. DATA ANALYSIS AND PROBLEM DESCRIPTION

This paper develops two comparative ANN models step-by-step to predict the stock price over financial time series, using data available at the website http://www.google.com/finance. The problem described in this paper is a predictive problem. In this paper four predictors have been used with one predictand. The four predictors are listed below

 Stock open price
 Stock price high
 Stock price low
 Stock close price
 Total trading volume

The predictand is next stock opening price.

All these four predictors of year X are used for prediction of stock opening price of year ( X+1). Whole dataset comprises of 1460 days NASDAQ stock data. Now first subset contains early 730 days data (open,high,low,close,volume) which is the inputseries to the neural network predictor.Second subset has later 730 days data(only open) which is the target series to the neural network predictor.Now the network learns the dynamic relationship between those previous five parameters (open, high, low, close, volume)to the one final parameter(open),which it will predict in future.

### A. Data Preprocessing

Once the historical stock prices are gathered ,now this is the time for data selection for training,testing and simulating the network.In this project we took 4 years historical price of any stock ,means total 1460 working days data.We done R/S analysis over these datafor predictability(Hurst exponent analysis).Now The Hurst exponent (H) is a statistical measure used to classify time series. H=0.5 indicates a random series while H>0.5 indicates a trend reinforcing series. The larger the H value is, the stronger trend. (1) H=0.5 indicates a random series. (2) 0<H<0.5 indicates an anti-persistent series. (3) 0.5<H<1 indicates a persistent series. An antipersistent series has a characteristic of "mean-reverting", which means an up value is more likely followed by a down value, and vice versa. The strength of "meanreverting" increases as H approaches 0.0. A persistent series is trend reinforcing, which means the direction (up or down compared to the last value) of the next value imore likely the same as current value. The strength of trend increases as H approaches 1.0. Most economic and financial time series are persistent with H>0.5. Now we took the dataset timeseries having hurst exponent >0.5 for persistency in good predictability.

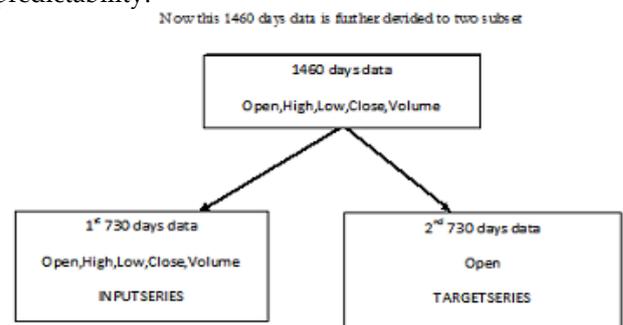

**Figure1. Data Division for NetworkTraining**

Now first subset contains early 730 days data(open,high,low,close,volume) which is the inputseries to the neural network predictor.Second subset has later 730 days data(only open) which is the target series to the neural network predictor.Now the network learns the dynamic relationship between those previous five parameters (open,high,low,close,volume) to the one final parameter(open),which it will predict in future.

All five predictors are given to the network & also corresponding predictand is given by using backpropagation traing (gradient descent approach) the network will learn the abstract mapping between input & output & will minimize prediction error.After getting satisfactory minimization of mean square error over several epoch the training is said to be completed & the prediction system is ready for forecasting purpose.

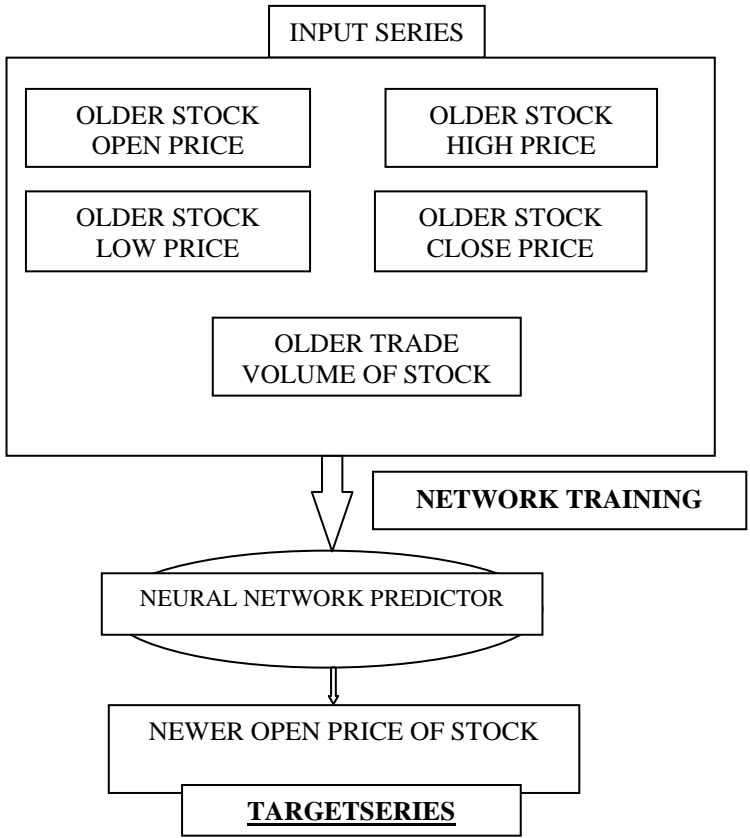

Figure2.Flow Chart for Data preprocessing & Training

After these data processing job is done these are fed to the network fortraining and testing,80% of total data is used for training purpose and rest 20% data is used for testing purpose.

## III. METHODOLOGY

This paper develops an ANN based comparative predictive model for NASDAQ stock prediction. The first ANN model is developed with Multi-Layer Feed forward Network Architecture & the second model is developed with Recurrent Neural Network Architecture. In this paper gradient descent based back propagation learning algorithm is used for the supervised learning of the predictive network. The mathematical model used in this paper is described below,

### B. Algorithm

Initialize each weight $w_i$ to some small random value.
- Until termination condition is met do ->
  -For each training example do ->
  - Input it & compute network output $O_k$
    - For each output unit k
      $\delta k \leftarrow O_k (1 - O)(T_k - O_k)$
    - For each hidden unit h
      $\delta h \leftarrow O_h (1 - O_h) \sum_{k \in outputs} w_{h,k} \delta k$
- For each network weight $w_i$ do –>
  $w_i \leftarrow w_i + \Delta w_{i,j}$
  Where $\Delta w_i = \eta \delta j x_i$,

Here the transfer function is sigmoid transfer function,it is used for its continuous nature. $\eta$ is the learning rate & is the gradient.
At first the network is constructed.in this paper ,sigmoidal function is used as the activation function of the ANN,it is chosen because of its continuous nature so the transfer function is eq(1),

$f(x) = (1 + e^{-x})^{-1}$ -------(1)

Where x is the total summed input received at node k. At first all weights are allocated to some small random value $w_i$ for ith layer. The successive weight is defined by eq(2),

$w_{i+1} = w_i + \Delta w$ -------(2)

The weight updating rule for gradient descent back propagation is eq(3),

$\Delta w_i = \eta \delta j x_i$, ----------(3)

Here we use mean square error,because the error surface is a multi-variable function it is wise to take mean of them & it is defined by eq(4),

$Err = \frac{1}{2}(Target\ Output - Network\ Output)^2$ ------------(4)

### III. IMPLEMENTATION AND RESULTS

The whole dataset is divided into training & test dataset,80% of total data is used for training purpose & 20% of total data is used for test purpose. Using gradient descent backpropagation algorithm the data are trained two times upto 1000 epochs. After training ANN model is tested over test dataset.Both networks are trained in same manner ,after completion of training comparison of their mean square error is presented by Table-1.

| Network Data | Feedforward NN | Timedelay Recurrent NN |
|---|---|---|
| Using Trainig Set | 4.14% | 3.01% |
| Using Testing Set | 25% | 15% |

Table-1 Comparison of ERROR.

A regression model relates $Y$ to a function of **X** and **β**.

$$Y \approx f(\mathbf{X}, \boldsymbol{\beta}) \quad \text{----------(5)}$$

The **unknown parameters** denoted as β; this may be a scalar or a vector.

The **independent variables**, **X**.

The **dependent variable**, $Y$.

Regression model is very much useful for model relation between function of independent variables & unknown parameters with some dependent variable. This paper also compute & contrast the regression plot for both networks over same NASDAQ data forecasting problem.

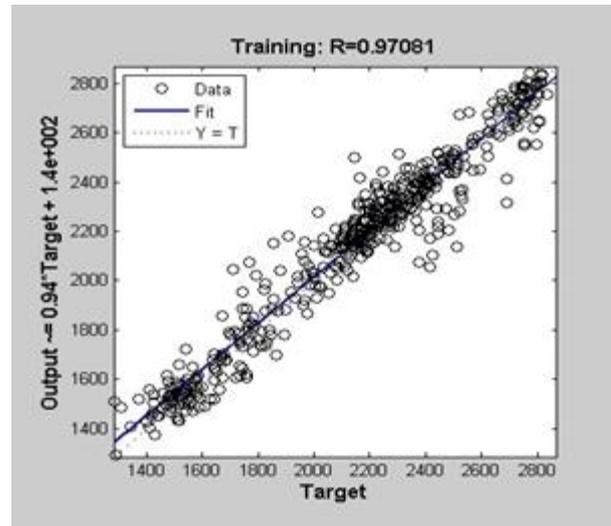

Figure 3. Regression plot for NASDAQ index (MLP)

Figure 3,depicts the regression plot for the feedforward MLP network, analyzing it we can say that Y=T regression is not so good.

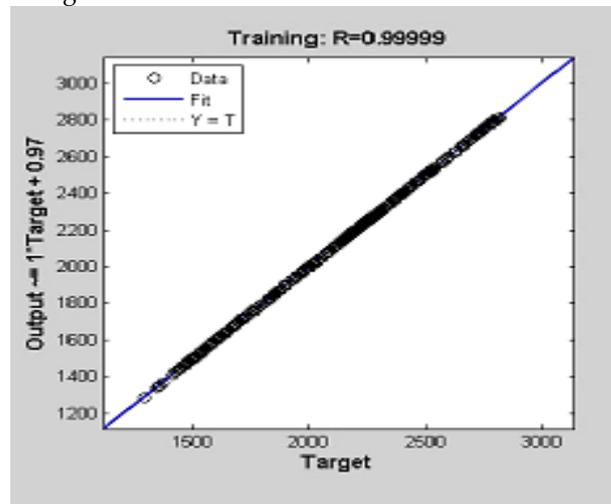

Figure 4: Regression plot for NASDAQ index (RNN)

Figure 4,depicts the regression plot for the Timedelay RNN network, analyzing it we can say that Y=T regression is totally fit.
This paper also comprises of comparative study of performance(mse) plot of both network.

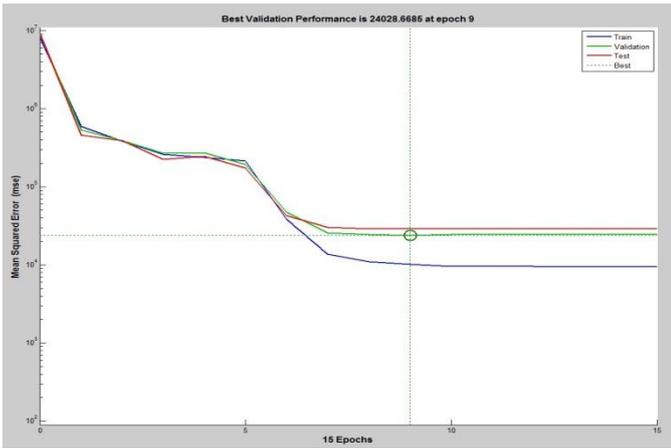

Figure 5: Performance plot for NASDAQ index (MLP)

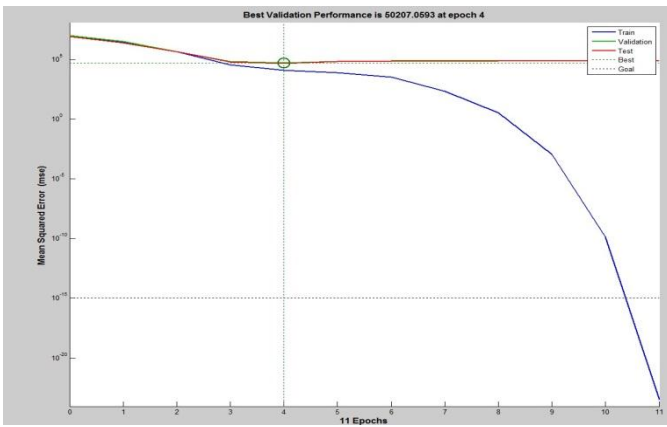

Figure 6: Performance plot for NASDAQ index (RNN)

Figure 5 we can see that the mse curve reaches the performance goal but it does not decrease in that good manner,but in Figure 6 the mse is reduces widely. By analyzing all these results one can say that RNN is better choice than Feedforward MLP in prediction purpose.

Table-2:Comparison between original stock price(TARGET) & Simulated price by ANN.
SIMRNN-simulated output using RNN model.
SIMMLP-simulated output using MLP model..

TABLE-2 REPRESENTS THE ORIGINAL STOCK VALUE & PREDICTED ONES

## IV. CONCLUSION

This paper presented a hybrid neural-evolutionary methodology to forecast time-series. The methodology is hybrid because an evolutionary computation-based

| TARGET | SIMRNN | SIMMLP | TARGET | SIMRNN | SIMMLP |
|---|---|---|---|---|---|
| 2519.1 | 2627 | 2183.8 | 2183.8 | 2627 | 2183.8 |
| 2514.8 | 5209.6 | 2379.7 | 2326.8 | 2352.8 | 2355.1 |
| 2500 | 3219 | 1921 | 2309.7 | 2355.3 | 2316.0 |
| 2473 | 3269.9 | 2215.3 | 2308.1 | 2353.4 | 2251.9 |
| 2509.4 | 2524.1 | 2083.9 | 2349.1 | 2416.2 | 2466.4 |
| 2528.5 | 2534.3 | 2534.3 | 2362.7 | 2420.3 | 2406.3 |
| 2483.2 | 2497.4 | 2124.7 | 2356.8 | 2356.8 | 2356.8 |
| 2436.7 | 2439.9 | 2141.6 | 2407.5 | 2456.4 | 2344.7 |
| 2444.9 | 2268.6 | 2309.9 | 2434.2 | 2412.8 | 2390.9 |
| 2414.4 | 2446.2 | 2086.0 | 2424.3 | 2521.9 | 2414.0 |
| 2423.0 | 2423.0 | 1961.4 | 2414.4 | 2462.8 | 2369.6 |
| 2443.1 | 2467.0 | 2018.2 | 2463.1 | 2463.1 | 2463.1 |
| 2481.2 | 2515.9 | 2239.5 | 2456.9 | 2363.3 | 2382.5 |
| 2444.9 | 2043.0 | 2223.8 | 2404.9 | 2501.4 | 2501.4 |
| 2416.5 | 2462.1 | 2107.2 | 2390.3 | 2438.9 | 2394.5 |
| 2375.8 | 2400.8 | 1992.4 | 2399.7 | 2435.9 | 2388.4 |
| 2388.4 | 2416.4 | 2227.8 | 2364.3 | 2888.2 | 2319.0 |
| 2365.8 | 2430.4 | 2119.0 | 2362.8 | 2412.8 | 2372.2 |
| 2319.6 | 2353.4 | 2173.1 | 2390.1 | 2444.5 | 2324.4 |
| 2312.4 | 2347.1 | 2120.1 | 2120.1 | 2441.5 | 2323.5 |
| 2274.2 | 2308.9 | 2102.5 | 2402.1 | 2468.7 | 2334.6 |
| 2311.5 | 2293.3 | 2242.3 | 2346.8 | 2421.8 | 2301.6 |
| 2261.7 | 2309.6 | 2251.3 | 2315.1 | 2315.1 | 2315.1 |
| 2263.6 | 2322.7 | 2089.0 | 2241.6 | 2312.9 | 2240.0 |
| 2244.9 | 2225.6 | 2337.0 | 2296.1 | 2369.7 | 2161.6 |
| 2290.6 | 2336.7 | 2204.6 | 2204.6 | 2204.6 | 2204.6 |
| 2239.9 | 2270.0 | 2102.5 | 2102.5 | 2102.5 | 2102.5 |
| 2102.5 | 2102.5 | 2463.1 | 2270.0 | 2463.1 | 2102.5 |
| 2456.9 | 2456.9 | 2270.0 | 2204.8 | 2102.5 | 2204.9 |
| 2308.9 | 2307.6 | 2308.9 | 2346.8 | 2102.5 | 2204 |
| 2362.8 | 2102.5 | 2390.1 | 2390.1 | 2270.0 | 2204 |
| 2362.8 | 2362.8 | 2270.0 | 2362.8 | 2390.1 | 2390.1 |
| 2362.8 | 2362.8 | 2364.3 | 2102.5 | 2270.0 | 2364.3 |
| 2501.4 | 2362.8 | 2270.0 | 2204 | 2204 | 2315.1 |
| 2362.8 | 2308.9 | 2089.0 | 2364.3 | 2315.1 | 2501.4 |
| 2501.4 | 2508.3 | 2362.8 | 2204 | 2102.5 | 2390.1 |

optimization process is used to produce a complete design of a neural network. The produced neural network, as a model, is then used to forecast the time-series. One of the advantages of the proposed scheme is that the design and training of the ANNs has been fully automated. This implies that the model identification does not require any human intervention. The model identification process involves data manipulation and a highly experienced statistician to do the work. This fact pushes the state of the art in automating the process of producing forecasting models. Compared to previous work, this paper approach is purely evolutionary, while others use mixed, mainly combined with back-propagation, which is known to get stuck in local optima. On the direction of model production, the evolutionary process automates the identification of input variables, allowing the user to avoid data pre-treatment and statistical analysis. The system is fully implemented in Matlab [15].

The study proves the nimbleness of ANN as a predictive tool for Financial Timeseries Prediction. Furthermore, Conjugate Gradient Descent is proved to be an efficient Backpropagation algorithm that can be adopted to predict the average stock price of NASDAQ.It is also revealed that temporal relationship between mapping is better learnt by RNN than FFMLP.

## ACKNOWLEDGMENT

Author heartily acknowledge Dr.Pabitra Mitra, Associate Professor Department Of Computer Science & Engineering,Indian Institute Of technology ,Kharagpur,India & Mr.Mriganka Chakraborty,Assistant Professor,Department of Computer Science & Engineering ,Seacom Engineering College,Howrah,India, for their endless help in this research work in theoretically & practically & specially thanks Prof.Rajob Bag,Head Of The Department,Department of Computer Science & Engineering ,Seacom Engineering College,Howrah,India, for his moral support. The design and simulation work was carried out at the laboratories of Computer Sciences Engineering at Seacom Engineering College ,Howrah India. Author must acknowledge the support of Seacom Engineering College authority in this paper publication.

## AUTHORS' VITAE

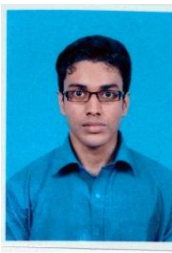

**Arka Ghosh** is currently pursuing Bachelor of Technology in Computer Science & Engineering from **Seacom Engineering College** under West Bengal University Of Technology West Bengal ,India. His research interest includes **Artificial Intelligence, Machine Learning, Networks ,Operating System& System Architecture.**